%% file: neurips_2019.tex
\documentclass{article}
\usepackage[dvipsnames]{xcolor}

\PassOptionsToPackage{numbers, compress}{natbib}

\usepackage[preprint]{neurips_2019}

\usepackage{sidecap} 
\sidecaptionvpos{figure}{c} 

\usepackage[utf8]{inputenc} 
\usepackage[T1]{fontenc}    
\usepackage{hyperref}       
\usepackage{url}            
\usepackage{booktabs}       
\usepackage{amsfonts}       
\usepackage{nicefrac}       
\usepackage{microtype}      
\usepackage{todonotes}
\input{header}

\usepackage[compact]{titlesec}
\titlespacing*{\section}{0pt}{0pt}{0pt}
\titlespacing*{\subsection}{0pt}{0pt}{0pt}

\AtBeginDocument{\setlength{\intextsep}{6pt}} 
\AtBeginDocument{\setlength{\textfloatsep}{6pt}} 
\AtBeginDocument{\setlength{\abovecaptionskip}{2pt}}
\AtBeginDocument{\setlength{\belowcaptionskip}{3pt}}
\AtBeginDocument{\setlength\abovedisplayskip{3pt}}
\AtBeginDocument{\setlength\belowdisplayskip{3pt}}

\graphicspath{{Figures/}}
\title{Information Geometry of\\Orthogonal Initializations and Training}

\author{%
  Piotr Aleksander Sok\'o\l{} and Il Memming Park \\
  Department of Neurobiology and Behavior\\
  Departments of Applied Mathematics and Statistics, and Electrical and Computer Engineering\\
  Institutes for Advanced Computing Science and AI-driven Discovery and Innovation\\
  Stony Brook University, Stony Brook, NY 11733 \\
  \texttt{\{memming.park, piotr.sokol\}@stonybrook.edu}
}

\begin{document}

\maketitle

\begin{abstract}
  Recently mean field theory has been successfully used to analyze properties of
  wide, random neural networks. It gave rise to a prescriptive theory for
  initializing feed-forward neural networks with orthogonal weights, which
  ensures that both the forward propagated activations and the backpropagated
  gradients are near $\ell_2$ isometries and as a consequence training is orders
  of magnitude faster. Despite strong empirical performance, the mechanisms by
  which critical initializations confer an advantage in the optimization of deep
  neural networks are poorly understood. Here we show a novel connection between
  the maximum curvature of the optimization landscape (gradient smoothness) as
  measured by the Fisher information matrix (FIM) and the spectral radius of the
  input-output Jacobian, which partially explains why more isometric networks
  can train much faster. Furthermore, given that orthogonal weights are
  necessary to ensure that gradient norms are approximately preserved at
  initialization, we experimentally investigate the benefits of maintaining
  orthogonality throughout training, from which we conclude that manifold
  optimization of weights performs well regardless of the smoothness of the
  gradients. Moreover, motivated by experimental results we show that a low
  condition number of the FIM is not predictive of faster learning.
\end{abstract}

\section{Introduction}
  Deep neural networks (DNN) have shown tremendous success in computer vision
  problems, speech recognition, amortized probabilistic inference, and the
  modelling of neural data. Despite their performance, DNNs face obstacles in
  their practical application, which stem from both the excessive computational
  cost of running gradient descent for a large number of epochs, as well as the
  inherent brittleness of gradient descent applied to very deep models. A number
  of heuristic approaches such as batch normalization, weight normalization and
  residual connections \cite{he_deep_2015,ioffe_batch_2015,salimans_weight_2016}
  have emerged in an attempt to address these trainability issues.

  Recently mean field theory has been successful in developing a more principled
  analysis of gradients of neural networks, and has become the basis for a new
  random initialization principle. The mean field approach postulates that in
  the limit of infinitely wide random weight matrices, the distribution of
  pre-activations converges weakly to an isotropic Gaussian. Using this
  approach, a series of works proposed to initialize the networks in such a way
  that for each layer the input-output Jacobian has mean singular values of 1
  \cite{schoenholz_deep_2016}. This requirement was further strengthened to
  suggest that the spectrum of singular values of the input-output Jacobian
  should concentrate on 1, and that this can only be achieved with random
  orthogonal weight matrices.

  Under these conditions the backpropagated gradients are bounded in $\ell_2$
  norm \cite{pennington_resurrecting_2017} irrespective of depth, i.e., they
  neither vanish nor explode. It was shown experimentally in
  \cite{pennington_resurrecting_2017,xiao_dynamical_2018-1,chen_dynamical_2018}
  that networks with these \textit{critical} initial conditions train orders of
  magnitude faster than networks with arbitrary initializations. The empirical
  success invites questions from an optimization perspective on how the spectrum
  of the hidden layer input-output Jacobian relates to notions of curvature of
  the parameters space, and subsequently to convergence rate. The largest
  effective initial step size $\eta_{0}$ is proportional to $ \frac{\vert m\vert
  }{M}$ for stochastic gradient descent, where the Hessian plays a central role
  for determining the local gradient smoothness $M$ and the strong convexity
  $m$\footnote{Recall that $m$ is the smallest, potentially negative eigenvalue
  of the Hessian and $M$ is its largest eigenvalue for twice differentiable
  objectives.} \cite{bottou_optimization_2016,boyd_convex_2004}. Recent attempts
  have been made to analyze the mean field geometry of the optimization
  landscape using the Fisher information matrix (FIM)
  \cite{amari_fisher_2018,karakida_universal_2018}, which given its close
  correspondence with the Hessian of the neural network defines an approximate
  gradient smoothness. \citet{karakida_universal_2018} derived an upper bound on
  the maximum eigenvalue, however this bound is not satisfactory since it is
  agnostic of the entire spectrum of singular values and therefore cannot
  differentiate between Gaussian and orthogonal initalizations.\\
  In this paper, we develop a new bound on the parameter
  space curvature \(M\) given the maximum eigenvalue of the Fisher information
  matrix $\lambda_{max}(\bar{\G})$ under both Gaussian and orthogonal
  initializations. We show that this quantity is proportional to the maximum
  squared singular value of the input-output Jacobian. We use this result to
  probe different orthogonal initializations, and observe that, broadly
  speaking, networks with a smaller initial curvature train faster and
  generalize better, as expected. However, consistently with a previous report
  \cite{pennington_emergence_2018}, we also observe highly isometric networks
  perform worse despite having a very small initial $\lambda_{max}(\bar{\G})$.
  We propose a theoretical explanation for this phenomenon using the connections
  between the FIM and the recently introduced Neural Tangent Kernel
  \cite{jacot_neural_2018, lee_wide_2019}. Given that the smallest and largest
  eigenvalues have an approximately inverse relationship
  \cite{karakida_universal_2018}, we propose an explanation that the long term
  optimization behavior is mostly controlled by the smallest eigenvalue \(m\)
  and therefore surprisingly there is a \emph{sweetspot with the condition
  number being \(\frac{m}{M} > 1\)}.\\ We then investigate whether constraining
  the spectrum of the Jacobian matrix of each layer affects optimization rate.
  We do so by training networks using Riemannian optimization to constrain their
  weights to be orthogonal, or nearly orthogonal and we find that manifold
  constrained networks are insensitive to the maximal curvature at the beginning
  of training unlike the unconstrained gradient descent (``Euclidean''). In
  particular, we observe that the advantage conferred by optimizing over
  manifolds cannot be explained by the improvement of the gradient smoothness as
  measured by $\lambda_{max}(\bar{\G})$, which argues against the proposed role
  of Batch Normalization recently put forward in
  \cite{santurkar_how_2018,yang_mean_2018}. Importantly, Euclidean training with
  a carefully designed initialization reduces the test misclassification loss at
  approximately the same rate as their manifold constrained counterparts, and
  overall attain a higher accuracy.

\section{Background}

\subsection{Formal Description of the Network}\label{sec:description}

  Following
  \citep{pennington_resurrecting_2017,pennington_emergence_2018,schoenholz_deep_2016},
  we consider a feed-forward, fully connected neural network with $L$ hidden
  layers. Each layer $l \in \{1,\dots,L \}$ is given as a recursion of the form
  \begin{equation} \x^l = \phi(\h^l), \quad \h^l = \W^{l}\x^{l-1} + \mathbf{b}^l
  \end{equation} where $\x^{l}$ are the activations, $\h^l$ are the
  pre-activations, $\W^l \in \RR^{N^{l} \times N^{l-1}}$ are the weight
  matrices, $\mathbf{b}^l$ are the bias vectors, and $\phi(\cdot)$ is the
  activation function. The input is denoted as $\x^0$. The output layer of the
  network computes $\mathbf{\hat{y}} = g^{-1}(\h^g) $ where $g$ is the link
  function and $\h^g = \W^g x^{L} + \mathbf{b}^g$.\\ The hidden layer
  input-output Jacobian matrix $\J^{\x^L}_{\x^0}$ is, \begin{equation}
  \J^{\x^L}_{\x^0} \defeq \frac{\partial \x^L}{\partial \x^0} =
  \prod^{L}_{l=1}\mathbf{D}^l\W^{l} \end{equation} where $\mathbf{D}^l$ is a
  diagonal matrix with entries $\mathbf{D}^l_{i,i} = \phi^\prime(\h^l_i)$. As
  pointed out in \citep{pennington_resurrecting_2017,schoenholz_deep_2016}, the
  conditioning of the Jacobian matrix affects the conditioning of the
  back-propagated gradients for all layers.

\subsection{Critical Initializations} Extending the classic result on the
Gaussian process limit for wide layer width obtained by
\citet{neal_bayesian_1996}, recent work
\citep{matthews_gaussian_2018,lee_deep_2017} has shown that for deep untrained
networks with elements of their weight matrices $\W_{i,j}$ drawn from a Gaussian
distribution $\calN(0,\frac{\sigma^2_\W}{N^l})$ the empirical distribution of
the pre-activations $\h^l$ converges weakly to a Gaussian distribution
$\calN(0,q^l\mathbf{I})$ for each layer $l$ in the limit of the width $N \to
\infty$. Similarly, it has been postulated that random orthogonal matrices
scaled by $\sigma_\W$ give rise to the same limit. Under this mean-field
condition, the variance of the pre-activation distribution $q^l$ is recursively
given by,
  \begin{equation} \label{eq:recursion} q^l = \sigma^2_\W \int
  \phi\left(\sqrt[]{q^{l-1}}h\right) \mathrm{d}\mu(h) + \sigma^2_\mathbf{b}
  \end{equation} where $\mu(h)$ denotes the standard Gaussian measure $\int
  \frac{\mathrm{d}h}{\sqrt[]{2\pi}}\exp{(\frac{-h^2}{2})}$ and
  $\sigma^2_\mathbf{b}$ denotes the variance of the Gaussian distributed biases
  \cite{schoenholz_deep_2016}. The variance of the first layer pre-activations
  $q^1$ depends on $\ell_2$ norm squared of inputs $q^1 =
  \frac{\sigma^2_\W}{N^1}\norm{\x^0}^2_2 + \sigma^2_\mathbf{b}$. The recursion
  defined in \eqref{eq:recursion} has a fixed point \begin{equation}
  \label{eq:fp} q^* = \sigma^2_\W \int \phi\left(\sqrt[]{q^{\ast}}h\right)
  \mathrm{d}\mu(h) + \sigma^2_\mathbf{b} \end{equation} which can be satisfied
  for all layers by appropriately choosing $\sigma_\W, \sigma_\mathbf{b}$ and
  scaling the input $\x^0$.
  To permit the mean field analysis of backpropagated signals, the authors
  \citep{schoenholz_deep_2016, pennington_resurrecting_2017,
  pennington_emergence_2018,karakida_universal_2018} further assume the
  propagated activations and back propagated gradients to be independent.
  Specifically, \begin{asu}\label{assumption}[Mean field assumptions]\\ \subasu
  \(\lim_{N \to \infty} \W \xrightarrow[]{d} \calN(0,q^*) \)
  \label{GaussianAssumption}\\ \subasu \(\lim_{N \to \infty} \cov \left[
  \J^g_{\x^{i+1}} \h^i , \J^g_{\x^{j+1}} \h^j \right] = 0  \quad \) for all \(
  i\neq j \) \label{IndependentAssumption} \end{asu} Under this assumption, the
  authors \cite{schoenholz_deep_2016,pennington_resurrecting_2017} analyze
  distributions of singular values of Jacobian matrices between different layers
  in terms of a small number of parameters, with the calculations of the
  backpropagated signals  proceeding in a selfsame fashion as calculations for
  the forward propagation of activations. The corollaries of Assumption
  \ref{assumption} and condition in \eqref{eq:fp} is that  $\phi^\prime(\h^l)$
  for $1\le l \le L$ are i.i.d. 
  In order to ensure that $\J^{\x^L}_{\x^0}$ is well conditioned,
  \citet{pennington_resurrecting_2017} require that in addition to the variance
  of pre-activation  being constant for all layers, two additional constraints
  be met. Firstly, they require that the mean square singular value of
  $\mathbf{D}\W$ for each layer have a certain value in expectation.
  \begin{equation} \label{eq:chi} \chi = \frac{1}{N}\EE\!\left[{}\Tr\!\left[
  (\mathbf{D}\W)\!{}\T\mathbf{D}\W \right]\right] = \sigma^2_\W \int\!\left[
  \phi^{\prime}(\sqrt[]{q^*}h) \right]^2 \mathrm{d}\mu(h) \end{equation} %
    Given that the mean squared singular value of the Jacobian matrix
    $\J^{\x^L}_{\x^0}$ is $(\chi)^L$, setting $\chi = 1$ corresponds to a
    critical initialization where the gradients are asymptotically stable as $L
    \to \infty$. Secondly, they require that the maximal squared singular value
    $s^2_{max}$ of the Jacobian $\J^{\x^L}_{\x^0}$ be bounded.
    \citet{pennington_resurrecting_2017} showed that for weights with Gaussian
    distributed elements, the maximal singular value increases linearly in depth
    even if the network is initialized with $\chi = 1$. Fortunately, for
    orthogonal weights, the maximal singular value $s_{max}$ is bounded even as
    $L \to \infty$ \citep{pennington_emergence_2018}.

    \section{Theoretical results: Relating the spectra of Jacobian and Fisher information matrices}\label{sec:Fisher}

    To better understand the geometry of the optimization landscape, we wish to put a Lipschitz bound on the gradient, which in turn gives an upper bound on the largest step size of any first order optimization algorithm.
    We seek to find local measures of curvature along the optimization trajectory.
    As we will show below the approximate gradient smoothness is tractable for random neural networks.
    The analytical study of Hessians of random neural networks started with \cite{pennington_geometry_nodate},
    but was limited to shallow architectures. Subsequent work \cite{amari_fisher_2018,karakida_universal_2018} on second order geometry of random networks shares
much of the spirit of the current work, in that it proposes to replace the
possibly indefinite Hessian with the related Fisher information matrix. The
Fisher information matrix plays a fundamental role in the geometry of
probabilistic models, under the Kullback-Leibler divergence loss. However,because of its relation to the Hessian, it can also be seen as defining an
approximate curvature matrix for second order optimization. Recall that the FIM
is defined as
     \begin{definition} Fisher Information Matrix
         \begin{align}
           \G & \defeq  \EE_{ \mathbf{y} \vert \x^0}\left[ \EE_{\x^0} \left[
           \nabla_{\mathbf{\theta}} \log{p_{\mathbf{\theta}}(\mathbf{y} | \x^{0})}
           \nabla_{\mathbf{\theta}} \log{p_{\mathbf{\theta}}(\mathbf{y} | \x^{0})} {}\T
           \right] \right] \label{eq:FishScore}\\
           & = \EE_{ \mathbf{y} \vert \x^0}\left[ \EE_{\x^0} \left[
            \J_{\theta}^{{h^{g}} \TT} \nabla^2_{h^g} \mathcal{L} \, \J_{\theta}^{h^g}
           \right]\right] 
	   = \EE_{ \mathbf{y} \vert \x^0}\left[ \EE_{\x^0} \left[
            \mathbf{H} - \sum_{k}{\nabla_{\x^g} \mathcal{L}_k \, \nabla^2_{\theta} h^g_k}
            \right] \right] \label{eq:HessToFisher}
         \end{align}
     \end{definition}
    where \(\mathcal{L}\) denotes the loss and \(\h^g\) is the output layer. The relation between the
 Hessian and Fisher Information matrices is apparent from
 \eqref{eq:HessToFisher}, showing that the Hessian \(\mathbf{H}\) is a quadratic
 form of the Jacobian matrices plus the possibly indefinite matrix of second
 derivatives with respect to parameters.

    Our goal is to express the gradient smoothness using the results of the previous section. Given \eqref{eq:HessToFisher} we can derive an analytical approximation to the Lipschitz bound using the results from the previous section; i.e. we will express the expected maximum eigenvalue of the random Fisher information matrix in terms of the expected maximum singular value of the Jacobian $\J^{\h^L}_{\h^1}$. To do so, let us consider the output of a multilayer perceptron as defining a conditional probability distribution $p_{\theta}(\mathbf{y} | \x^{0})$, where $\Theta = \{\vec{\W^1},\ldots,\vec{\W^L} , \mathbf{b}^1, \ldots, \mathbf{b}^L\}$ is the set of all hidden layer parameters, and $\theta$ is the column vector containing the concatenation of all the parameters in $\Theta$.
    Then each random block of the Fisher information matrix with respect to parameter vectors $a,b \in \Theta$ can further be expressed as
    \begin{equation} \label{eq:BlockJac}
         \bar{\G}_{a,b} =
         \J^{\h^{g}}_{a} {}\T
        \mathbf{H}_g
         \J^{\h^{g}}_{b }
    \end{equation}
  where the final layer Hessian $\mathbf{H}_g$ is defined as $\nabla^2_{\h^g}
  \log p_\theta(\mathbf{y}|\x^0)$. We can re-express the outer product of the
  score function $\nabla_{\h^g} \log p_\theta(\mathbf{y}|\x^0)$  as the second
  derivative of the log-likelihood (see \eqref{eq:FishScore}), provided it is
  twice differentiable and it does not depend on $\mathbf{y}$, which also allows
  us to drop conditional expectation with respect to $\mathbf{y} \vert \x^0$.
  This condition naturally holds for all canonical link functions and matching
  generalized linear model loss functions. We define the matrix of partial
  derivatives of the $\alpha$-th layer pre-activations with respect to the layer
  specific parameters separately for $\W^\alpha$ and $\mathbf{b}^\alpha$ as:
  \begin{align}\label{eq:JacParam} \J^{\h^{\alpha}}_{a} &= \x^{\alpha - 1} {}\T
  \otimes \I &\quad \text{for }& a = \vec{\W^\alpha}\\ \J^{\h^{\alpha}}_{a} &=
  \I &\quad \text{for }& a = \mathbf{b}^\alpha \end{align} Under the assumptions
  in \ref{assumption}, we can further simplify the expression for the blocks of
  the Fisher information matrix \eqref{eq:BlockJac}.
\begin{lemma}\label{lem:BlockWeights} The expected blocks with respect to weight
matrices for all layers \(\alpha, \beta \neq 1 \) are
 \begin{align} \bar{\G}_{\vec{\W^\alpha}, \vec{\W^\beta}} & = %
\EE \left[ \x^{\alpha - 1}\x^{\beta- 1} {}\T \right] \otimes
\J^{\h^g}_{\h^{\alpha}} {}\T \mathbf{H}_g \J^{\h^g}_{\h^{\beta}} \\ 
\left[ 
\J^{\h^g}_{\h^{\alpha}} {}\T 
\right] 
\EE\left[ 
\ones {\ones}^\TT 
\J^{\h^g}_{\h^{\alpha}} {}\T 
\right] \label{eq:BlockWeights} \end{align} \end{lemma} \begin{lemma}
\label{lem:CrossBlock} The expected blocks with respect to a weight matrix
 \(\W^\alpha\) and a bias vector \(\mathbf{b}^\beta\)  are
\begin{equation} \bar{\G}_{\vec{\W^\alpha}, \mathbf{b}^\beta}=  \\ \left[
\x^{\alpha - 1} {}\T \otimes \I \, \right] \J^{\h^g}_{\h^{\alpha}} {}\T
\mathbf{H}_g \J^{\h^g}_{\h^{\beta}}
    \end{equation}
  \end{lemma}

Leveraging lemmas \ref{lem:BlockWeights} and \ref{lem:CrossBlock}, and the
previously derived spectral distribution for the singular values of the
Jacobians, we derive a block diagonal approximation which in turn allows us to
bound the maximum eigenvalue \(\lambda_{max}(\bar{\G})\). In doing so we will
use a corollary a of the block Gershgorin theorem.
\begin{proposition}[(informal) Block Gershgorin theorem] \label{thm:GRSHG} The
maximum eigenvalue \(\lambda_{max}(\bar{\G})\) is contained in a union of disks
centered around the maximal eigenvalue of each diagonal block with radia equal
to the sum of the singular values of the off-diagonal terms. \end{proposition}
For a more formal statement see Appendix~\ref{app:GRSHG}. The proposition
\ref{thm:GRSHG} suggest a simple, easily computable way to bound the expected
maximal eigenvalue of the Fisher information matrix---choose the block with the
largest eigenvalue and calculate the expectedspectral radia for the
corresponding off diagonal terms. We do so by making an auxiliary assumption:
\begin{asu} The maximum singular value of \(\J^{\h^g}_{\h^\alpha}\)
monotonically increases as \(\alpha \downarrow 1\). \end{asu} \begin{figure}[t]
\centering \includegraphics[width=\linewidth]{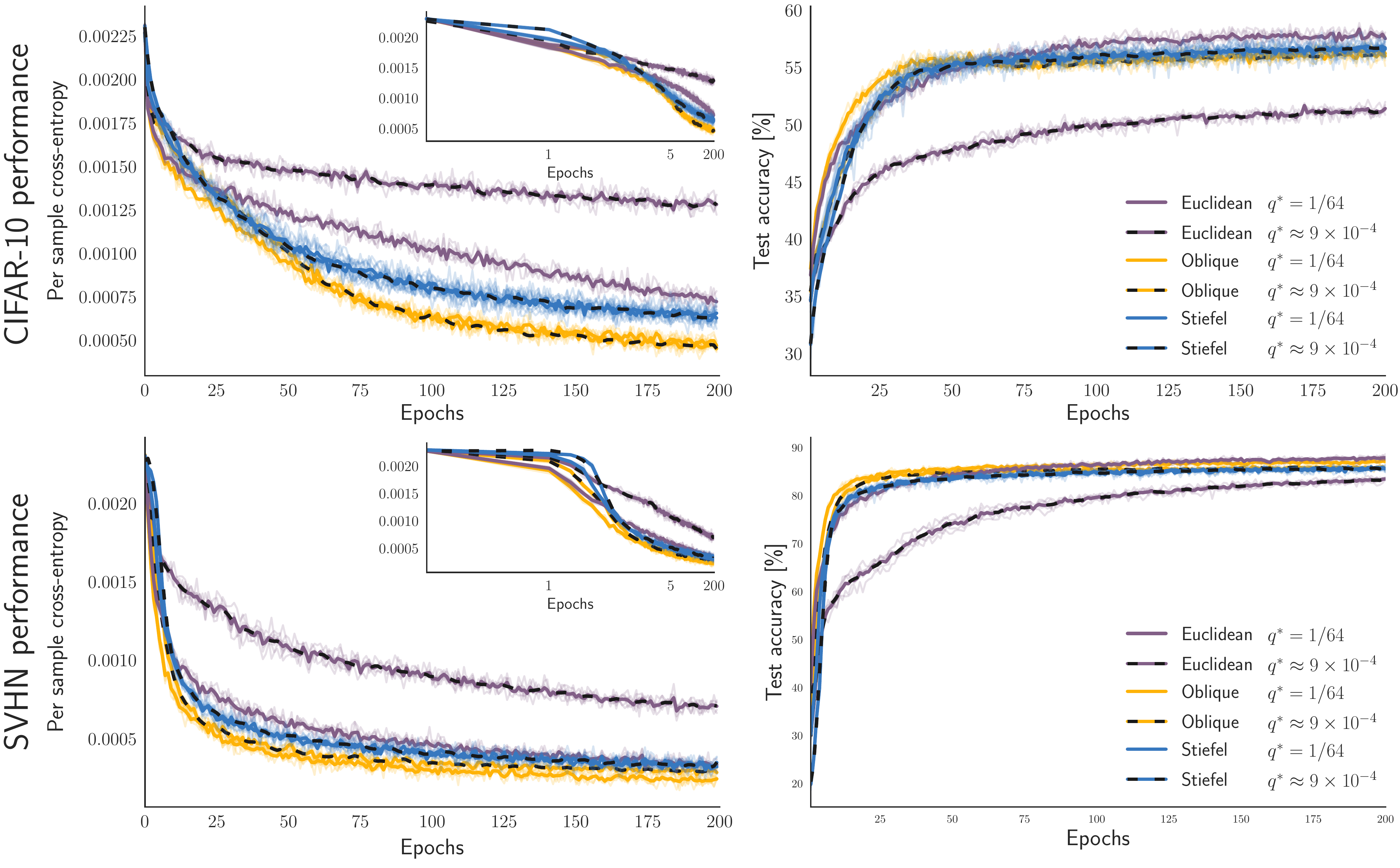}
\caption{ \textbf{Manifold constrained networks are insensitive to the choice of
$q^*$}: Train loss and test accuracy for Euclidean, Stiefel and Oblique networks
with two different values of $q^*$. The manifold constrained networks minimize
the training loss at approximately the same rate, being faster than both
Euclidean networks. Despite this, there is little difference between the test
accuracy of the Stiefel and Oblique networks and the Euclidean networks
initialized with $q^* = 9 \times 10^{-4}$. Notably, the latter attains a
marginally higher test set accuracy towards the end of training. }
\label{fig:losses} \end{figure}[t] The assumption that the maximal singular of
the Jacobians \(\J^{\h^g}_{\h^\alpha}\) grows with backpropagated depth is well
supported by previous observations
\cite{pennington_resurrecting_2017,pennington_emergence_2018}. Under this
condition it is sufficient to study the maximal singular value of blocks of the
Fisher information matrix with respect to \(\vec{\W^1},b^1\) and the spectral
norms of its corresponding off-diagonal blocks. We define functions
\(\textcolor{PineGreen}{\Sigma_{max}}\) of each block as upper bounds on the spectral bounds of the
respective block. The specific values are given in the following Lemma \ref{lem:Bounds}.

  \begin{lemma}\label{lem:Bounds} The maximum expected singular values
\(\EE\left[\sigma_{max}\right]\) of the off-diagonal blocks \(\forall \beta \neq
1\) are bounded by \(\textcolor{PineGreen}{\Sigma_{max}}\):\\

\begin{flalign}
\sigma_{max}\left({\G}_{\vec{\W^1}, \vec{\W^\beta}}\right) \le
\textcolor{PineGreen}{\Sigma_{max}\left({\G}_{\vec{\W^1}, \vec{\W^\beta}}\right)}
 &\\
 \defeq  \textcolor{PineGreen}{  \sqrt{N^\beta} \left\vert  \EE\left[\phi(h)\right] \right\vert
\left\lVert\EE\left[\x^0\right]\right\rVert_2 \EE\left[
\sigma_{max}\left(\J^{\h^g\TT}_{\h^1}\right) \right] \EE\left[
\sigma_{max}\left( \mathbf{H}_g \right) \right] \EE\left[ \sigma_{max}\left(
\J^{\h^g}_{\h^\beta} \right] \right) } & \end{flalign}

\begin{align}
\sigma_{max}\left({\G}_{\vec{\W^1}, {b^\beta}}\right) &\le
\textcolor{PineGreen}{\Sigma_{max}\left({\G}_{\vec{\W^1}, {b^\beta}}\right)} & \\ &\defeq
\textcolor{PineGreen}{\left\vert
\EE\left[\phi(h)\right] \right\vert \EE\left[
\sigma_{max}\left(\J^{\h^g\TT}_{\h^1}\right) \right] \EE\left[
\sigma_{max}\left( \mathbf{H}_g \right) \right] \EE\left[ \sigma_{max}\left(
\J^{\h^g}_{\h^\beta} \right) \right]} \end{align}
\begin{align}
\sigma_{max}\left({\G}_{b^1, b^\beta}\right) \le \textcolor{PineGreen}{\Sigma_{max}\left({\G}_{b^1,
b^\beta}\right)} \defeq \textcolor{PineGreen}{\EE\left[  \sigma_{max}\left(\J^{\h^g\TT}_{\h^1}\right)
\right] \EE\left[ \sigma_{max}\left( \mathbf{H}_g \right) \right] \EE\left[
\sigma_{max}\left( \J^{\h^g}_{\h^\beta} \right) \right]} \end{align}
\begin{proof} See Appendix \ref{app:SingVals}
    \end{proof}
  \end{lemma}
  Note that the expectations for layers \(> 1\) is over random networks
  realizations and averaged over data \(\x^0\); i.e. they are taken with respect
  to the Gaussian measure, whereas the expectation for first layer weights is
  taken with respect to the empirical distribution of \(\x^0\) (see
  \eqref{eq:fp}). \begin{lemma} The maximal singular values of the block
  diagonal elments are bounded by \(\textcolor{PineGreen}{\Sigma_{max}}\) \begin{align}
  \sigma_{max}\left({\G}_{b^1,b^1}\right) &\le
  \textcolor{PineGreen}{\Sigma_{max}\left({\G}_{b^1,b^1}\right)} \\ &\defeq
  \textcolor{PineGreen}{\EE\left[
  \sigma_{max}\left( \mathbf{H}_g \right) \right] \EE\left[ \sigma_{max}\left(
  \J^{\h^g}_{\h^\beta} \right) \right]^2} \end{align} \begin{align}
  \sigma_{max}\left({\G}_{\W^1,\W^1}\right) &\le
  \textcolor{PineGreen}{\Sigma_{max}\left({\G}_{\W^1,\W^1}\right)} \\ &\defeq
  \textcolor{PineGreen}{
  \sigma_{max}\left(\cov[\x^0,\x^0]\right) \EE\left[ \sigma_{max}\left(
  \mathbf{H}_g \right) \right] \EE\left[ \sigma_{max}\left( \J^{\h^g}_{\h^\beta}
  \right) \right]^2}
    \end{align}
    \end{lemma}

  Depending on the choice of $q^*$ and therefore implicitly both the rescaling
  of $\x^0$ and the values of $\EE[\phi(\h)]$ either of the quantities might
  dominate.

  \begin{theorem}[Bound on the Fisher Information Eigenvalues] \label{thm:Bound}
  If \( \left\lVert\EE\left[\x^0\right]\right\rVert_2 \le 1\) then eigenvalue
  associated with $b^1$ will dominate, giving an upper bound on
  $\lambda_{max}(\bar{G})$
  \begin{equation} \lambda_{max}({G}) \le
  \textcolor{PineGreen}{\Sigma_{max}\left({\G}_{{b^1}, {b^1}}\right)} + \sum_{\beta>1}
  \textcolor{PineGreen}{\Sigma_{max}\left({\G}_{{b^1}, {b^\beta}}\right)} +
  \textcolor{PineGreen}{\Sigma_{max}\left({\G}_{\vec{b^1}, \vec{\W^\beta}}\right)}\nonumber
  \end{equation} otherwise the maximal eigenvalue of the FIM is bounded by
    \begin{equation}
      \lambda_{max}({G}) \le \textcolor{PineGreen}{\Sigma_{max}\left({\G}_{\vec{\W^1}, \vec{\W^1}}\right)} +
       \sum_{\beta > 1}
      \textcolor{PineGreen}{\Sigma_{max}\left({\G}_{\vec{\W^1},b^\beta}\right)} +
      \textcolor{PineGreen}{\Sigma_{max}\left({\G}_{\vec{\W^1}, \vec{\W^\beta}}\right)}
      \nonumber
    \end{equation}
  \end{theorem}
  The functional form of the bound is essentially quadratic in \( \sigma_{max}(\J^{\h^g}_{\h^1}\) since the term appears in the summand as with powers at most two.
  This result shows that the strong smoothness, given by the maximum eigenvalue of the FIM, is \emph{proportional to} the squared maximum singular value of the input-output Jacobian \(\sigma_{max}\left(\EE\left[\J^{\h^g}_{\h^1}\right]\right)\).
  Moreover, the bound essentially depends on $q^*$ via the expectation $\EE[\phi(h)]$, through $\J^{\h^g}_{\h^1}$ and implicitly through $\mathbf{H}_g$. For regression problems this dependence is monotonically increasing in $q^*$ \cite{pennington_emergence_2018,pennington_resurrecting_2017} since $\mathbf{H}_g$ is just the identity. However, this does not hold for all generalized linear models since $\lambda_{max}(\mathbf{H}_g)$ is not necessarily a monotonically increasing function of the pre-activation variance at layer $\h^g$. We demonstrate this in the case of softmax regression in the Appendix ~\ref{app:Hg}.
  Finally, to obtain a specific bound on $\lambda_{max}(\bar{G})$ we might consider bounding each $\sigma_{max}(\EE\left[\J^{\h^g}_{\h^\alpha}\right])$ appearing in theorem \ref{thm:Bound} in terms of its Frobenius norm. The corresponding result is the eigenvalue bound derived by \citep{karakida_universal_2018}.
\section{Numerical experiments}
  \subsection{Manifold optimization}
    Next we test the role that maintaining orthogonality throughout training has on the optimization performance. Moreover we numerically probe our predictions concerning the proportionality between the maximal eigenvalues of the Fisher information matrix and the maximal singular values of the Jacobian. Finally we measure the behavior of \(\lambda_{max}(\bar{G})\) during training. To achieve this we perform optimization over manifolds.\\
   Optimizing neural network weights subject to manifold constraints has recently attracted considerable interest \citep{arjovsky_unitary_2015,henaff_recurrent_2016,vorontsov_orthogonality_2017, xie_all_2017,cho_riemannian_2017,ozay_optimization_2016, cisse_parseval_2017}. In this work we probe how constraining the weights of each layer to be orthogonal or near orthogonal affects the spectrum of the hidden layer input-output Jacobian and of the Fisher information matrix.
   In Appendix \ref{app:Manopt} we provide a review notions from differential geometry and optimization over matrix manifolds \citep{edelman_geometry_1998,absil_optimization_2007}.
   The Stiefel manifold and the oblique manifold will be used in the subsequent sections.
   \paragraph{Stiefel Manifold} \(\St(p,n) \defeq \{ \W \in \RR^{n\times p} : \W {}\T\W = \I _{p} \}\)
   \paragraph{Oblique Manifold} \( \Ob(p,n) \defeq \{ \W \in \RR^{n\times p} : \mathrm{diag}(\W {}\T\W) = \mathbf{1} \} \)\\
   Constraining the weights to this manifold is equivalent to using Weight Normalization \cite{salimans_weight_2016}.
   \citet{cho_riemannian_2017} derived a regularization term which penalizes the distance between the point in the manifold $\W$ and the closest orthogonal matrix with respect to the Frobenius norm.
  \begin{equation} \label{eq:orthopenalty}
      \rho(\lambda,\W) = \frac{\lambda}{2} \mnorm{\W {}\T\W - \mathbf{I}}^2_{F}
  \end{equation}

\subsection{Numerical Experiments}
  \begin{SCfigure}[45][t]
      \centering
     \includegraphics[width=0.65\linewidth]{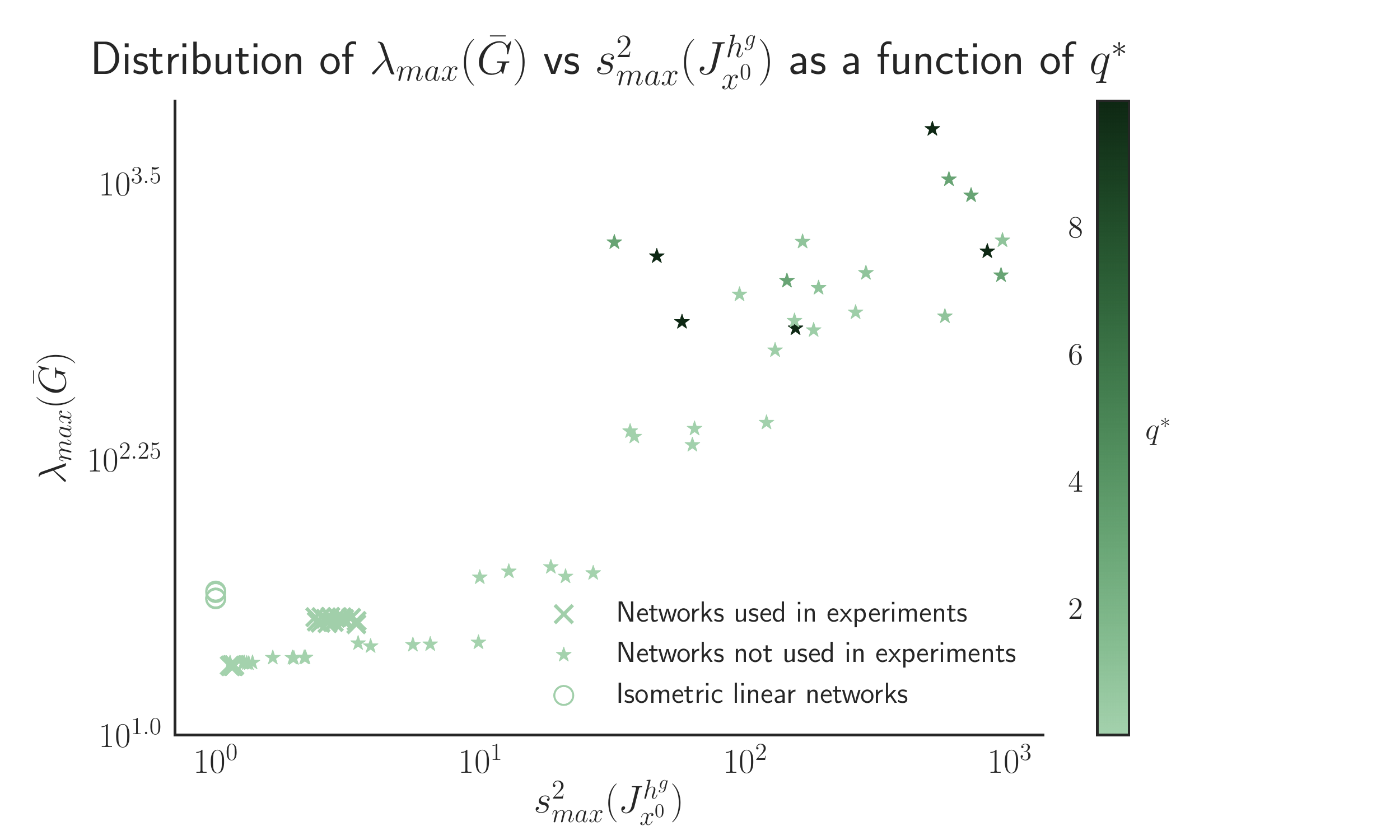}
     \caption{
         \textbf{At initialization the maximum eigenvalue of the Fisher information matrix $\bar{\G}$ correlates highly with the maximum squared singular value of the Jacobian $\J^{\mathbf{\h}^g}_{\mathbf{\x}^0}$ (\(\rho=0.65\)).}
     }
     \label{fig:preactivations:lambdaSigmaA}
  \end{SCfigure}
  To experimentally test the potential effect of maintaining orthogonality throughout training and compare it to the unconstrained optimization~\citep{pennington_resurrecting_2017}, we trained a 200 layer $\mathrm{tanh}$ network on CIFAR-10 and SVHN. Following \citep{pennington_resurrecting_2017} we set the width of each layer to be $N= 400$ and chose the $\sigma_\W, \, \sigma_\mathbf{b}$ in such a way to ensure that $\chi$ concentrates on 1 but $s_{max}^2$ varies as a function of $q^*$ (see Fig.~\ref{fig:preactivations:lambdaSigmaA}). We considered two different critical initializations with $q^* = \frac{1}{64}$ and $q^* \approx 9\times 10^{-4}$, which differ both in spread of the singular values as well as in the resulting training speed and final test accuracy as reported by~\citep{pennington_resurrecting_2017}.
  To test how enforcing strict orthogonality or near orthogonality affects convergence speed and the maximum eigenvalues of the Fisher information matrix, we trained Stiefel and Oblique constrained networks and compared them to the unconstrained ``Euclidean'' network described in \citep{pennington_resurrecting_2017}. We used a Riemannian version of ADAM~\citep{kingma_adam:_2014}. When performing gradient descent on non-Euclidean manifolds, we split the variables into three groups:
 (1) Euclidean variables (e.g. the weights of the classifier layer, biases),
 (2) non-negative scaling $\sigma_\W$ both optimized using the regular version of ADAM,
 and (3) manifold variables optimized using Riemannian ADAM.
 The initial learning rates for all the groups, as well as the non-orthogonality penalty (see \ref{eq:orthopenalty}) for Oblique networks were chosen via Bayesian optimization, maximizing validation set accuracy after 50 epochs.
 All networks were trained with a minibatch size of 1000. We trained 5 networks of each kind, and collected eigenvalue and singular value statistics every 5 epochs, from the first to the fiftieth, and then after the hundredth and two hundredth epochs.

   Based on the bound on the maximum eigenvalue of the Fisher information matrix derived in Section \ref{sec:Fisher}, we predicted that at initialization $\lambda_{max}(\bar{\G})$ should covary with $\sigma^2_{max}(\J^{\h^g}_{\x^0})$. Our prediction is vindicated in that we find a strong, significant correlation between the two (Pearson coefficient $\rho=0.64$).
   The numerical values are presented in Fig.~\ref{fig:preactivations:lambdaSigmaA}.
   Additionally we see that both the maximum singular value and maximum eigenvalue increase monotonically as a function of $q^*$.
   Motivated by the previous work by \citet{saxe_exact_2013} showing depth independent learning dynamics in linear orthogonal networks, we included 5 instantiations of this model in the comparison. The input to the linear network was normalized the same way as the critical, non-linear networks with $q^* = 1/64$. The deep linear networks had a substantially larger $\lambda_{max}(\bar{\G})$ than its non-linear counterparts initialized with identically scaled input (Fig.~\ref{fig:preactivations:lambdaSigmaA}).
   Having established a connection between $q^*$ the maximum singular value of the hidden layer input-output Jacobian and the maximum eigenvalue of the Fisher information, we investigate the effects of initialization on subsequent optimization.
   As reported by \citet{pennington_resurrecting_2017}, the learning speed and generalization peak at intermediate values of $q^* \approx 10^{-0.5}$. This result is counter intuitive given that the maximum eigenvalue of the Fisher information matrix, much like that of the Hessian in convex optimization, upper bounds the maximal learning rate  \citep{boyd_convex_2004,bottou_optimization_2016}.
   To gain insight into the effects of the choice of $q^*$ on the convergence rate, we trained the Euclidean networks and estimated the local values of $\lambda_{max}$ during optimization.
   At the same time we asked whether we can effectively control the two aforesaid quantities by constraining the weights of each layer to be orthogonal or near orthogonal. To this end we trained Stiefel and Oblique networks and recorded the same statistics.

       We present training results in Fig. \ref{fig:losses}, where it can be seen that Euclidean networks with $q^* \approx 9\times 10^{-4}$ perform worse with respect to training loss and test accuracy than those initialized with $q^* = 1/64$. On the other hand, manifold constrained networks are insensitive to the choice of $q^*$. Moreover, Stiefel and Oblique networks perform marginally worse on the test set compared to the Euclidean network with $q^* = 1/64$, despite attaining a lower training loss. This latter fact indicates that manifold constrained networks the are perhaps prone to overfitting.

       We observe that reduced performance of Euclidean networks initialized with $q^* \approx 9\times10^{-4}$ may partially be explained by their rapid increase in $\lambda_{max}(\bar{\G})$ within the initial 5 epochs of optimization (see Fig. \ref{fig:GradientSmoothness} in the Appendix).
       While all networks undergo this rapid increase, it is most pronounced for Euclidean networks with $q^* \approx 9\times10^{-4}$. The increase $\lambda_{max}(\bar{\G})$ correlates with the inflection point in the training loss curve that can be seen in the inset of Fig.~\ref{fig:losses}.
       Interestingly, the manifold constrained networks optimize efficiently despite differences in $\lambda_{max}(\bar{\G})$, showing that their performance cannot be attributed to increasing the gradient smoothness as postulated by \citep{santurkar_how_2018}. These results instead bolster support for the theory proposed by \cite{kohler_exponential_2018}
       \begin{figure}[t!bh]
        \centering
               \includegraphics[width=\linewidth]{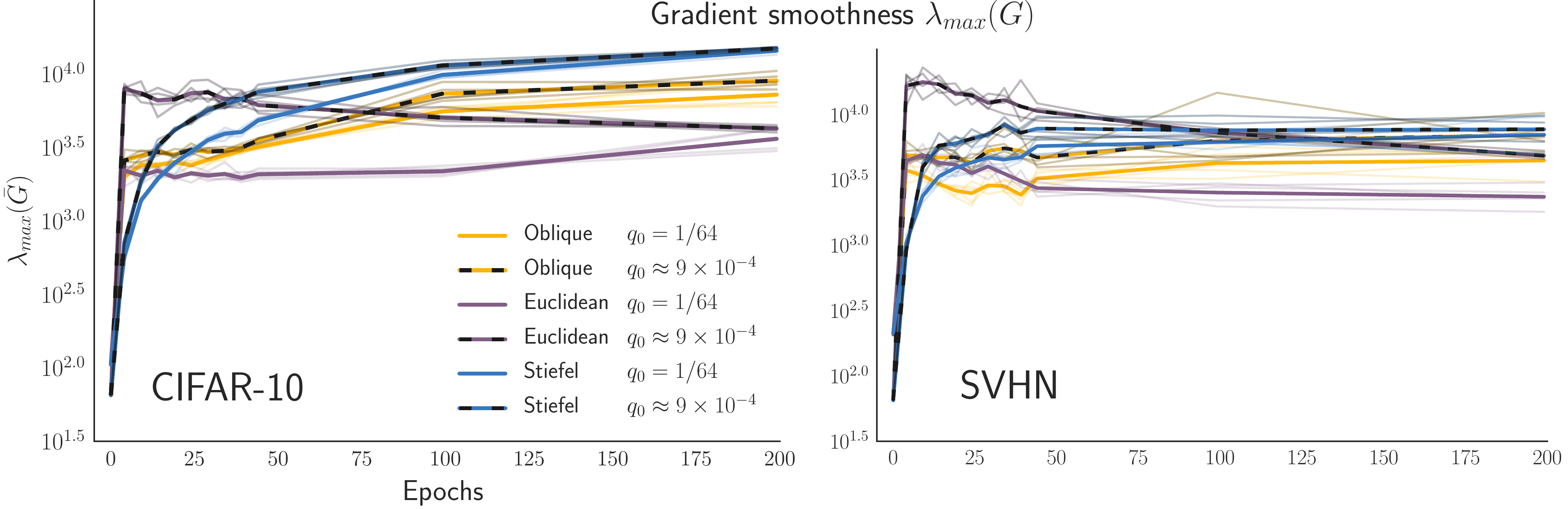}
           \caption{
               \textbf{For manifold constrained networks, gradient smoothness is not predictive of optimization rate. Euclidean networks with a low initial $\lambda_{max}(\bar{G})$ rapidly become less smooth, whereas Euclidean networks with a larger $\lambda_{max}(\bar{G})$ remain relatively smoother.
               }
               Notably, the Euclidean network with $q^* = 1/64$ has almost an order of magnitude smaller $\lambda_{max}(\bar{\G})$ than the Stiefel and Oblique networks, but reduces training loss at a slower rate.
           }
           \label{fig:GradientSmoothness}
       \end{figure}
       \section{Discussion}
       Critical orthogonal initializations have proven tremendously successful in rapidly training very deep neural networks \cite{pennington_resurrecting_2017,chen_dynamical_2018,pennington_emergence_2018,xiao_dynamical_2018}. Despite their elegant derivation drawing on methods from free probability and mean field theory, they did not offer a clear optimization perspective on the mechanisms driving their success. With this work we complement the understanding of critical orthogonal initializations by showing that the maximum eigenvalue of the Fisher information matrix, and consequentially the local gradient smoothness is proportional to the maximum singular value of the input-output Jacobian. This gives an information geometric account of why the step size and training speed depend on $q^*$ via its effect on $s_{max}(\EE\left[\J^{\h^L}_{\h^1}\right])$.
       We observed in numerical experiments that the paradoxical results reported in \cite{pennington_resurrecting_2017} whereby training speed and generalization attains a maximum for $q^* = 10^{-0.5}$ can potentially be explained by a rapid increase of the maximum eigenvalue of the FIM during training for the networks initialized with Jacobians closer to being isometric (i.e., smaller $q^\ast$).
       This increase effectively limits the learning rate during the early phase of optimization and highlights the need to analyze the trajectories of training rather than just initializations. We relate that to the recently proposed Neural Tangent Kernel\cite{jacot_neural_2018,lee_wide_2019}. The NTK is defined as
         \begin{equation} \label{eq:NTK}
           \hat{\Theta}_{t,i,j} \defeq \J^{\h^g}_{\x^0} \J^{\h^g \TT}_{\x^0}
         \end{equation}
         for $i,j \in N^g \lvert \calD \rvert$ representing the block indices running over $N^g$ outputs of the network and $\lvert \calD \rvert$ data samples. The NTK is the derivative of a kernel defined by a random neural network. It prescribes the time evolution of the function and therefore offers a concise description of the network predictions. Importantly, the spectrum of the NTK coincides with that of the Fisher information \emph{for regression problems} (see Appendix \ref{NTKJAC}).\\
         It is therefore interesting to understand the predictiveness of the Neural Tangent Kernel at initialization given its spectrum. Such a result has been recently presented by \citep{lee_wide_2019}, who show that the discrepancy between training with a NTK frozen at initialization ($f^{lin}_t(\x^0)$) and a continuously updated one ($f_t(,\x^0)$) can be bounded. Importantly the authors showed that rate at which discrepancy accrues depends exponentially on the smallest eigenvalue of the NTK. Given that the spectra of the Neural Tangent Kernel and the Fisher Information matrix coincide we can reason about this discrepancy over training time in terms of the smallest and largest eigenvalues of the Fisher Information matrix.
          \begin{lemma}[\citet{lee_wide_2019}] The discrepancy between $g^{lin}(t) = f_t^{lin}(\x^0) - \mathbf{y}$ and $g(t) = f_t(\x^0) - \mathbf{y}$
            \begin{equation}
               e ^ { \lambda _ {min }(\bar{\G}_0) \eta t } \left\| g ^ { \operatorname { lin } } ( t ) - g ( t ) \right\| _ { 2 } \le \\
                \left( \eta \int _ { 0 } ^ { t } e ^ { \lambda _ { min }(\bar{\G})_0 \eta s } \left\| \left( \bar{\G} _ { s } - \bar{\G} _ { 0 } \right) \right\| \left\| g ^ { \operatorname { lin } } ( s ) \right\| _ { 2 } d s  \right) e^ { \int_0^t {\left( \eta \left\| \left( \bar{\G} _ { s } - \bar{\G}_ { 0 } \right) \right\| \right) ds} }
            \end{equation}
          where $\eta$ is the learning rate.
        \end{lemma}
        Given the approximately inverse relation between the maximum and minimum eigenvalues of the Fisher information matrix \cite{karakida_universal_2018}, decreasing $q*$  increases \(\lambda _ {min }(\bar{\G}_0)\) and the the solutions rapidly diverge. This implies that a low condition number $\frac{\lambda _ {max}(\bar{\G}_0)}{\lambda _ {max}(\bar{\G}_0)}$ may be undesirable, and a degree of anisotropy is necessary for the Fisher Information matrix to be predictive of training performance.\\
       Finally, we compared manifold constrained networks with the Euclidean network, each evaluated with two initial values of $q^*$. From these experiments we draw the conclusion that manifold constrained networks are less sensitive to the initial strong smoothness, unlike their Euclidean counterparts. Furthermore, we observe that the rate at which Stiefel and Oblique networks decrease training loss is not dependent on their gradient smoothness, a result which is consistent with the recent analysis of \cite{kohler_exponential_2018}.

\bibliography{references}
\bibliographystyle{plainnat}

\clearpage
\section{Appendix}
\subsection{Block Gershgorin Theorem}\label{app:GRSHG}
    In Section \ref{sec:Fisher}, we considered a block diagonal approximation to the Fisher information matrix and derived an upper bound on the spectral norm for all the blocks. Using the properties of the off-diagonal blocks, we can get a more accurate estimate of the maximal eigenvalue of the Fisher information might be. First, let us consider an arbitrarily partitioned matrix $\A \in \RR^{N\times N}$, with spectrum $\lambda(\A)$ The partitioning is done with respect to the set
    \begin{equation}
        \pi =  \{ p_j\}^{L}_{j=0}
    \end{equation}
    with the elements of the set satisfying $0 < p_1 < p_2 < \ldots < p_L = N$.
    Then each block of the matrix $A_{i,j}$ is a potentially rectangular matrix in $\RR^{(p_i - p_{i-1})\times(p_j - p_{j-1})}$. We assume that $\A_{i,i}$ is self-adjoint for all $i$.\\
    Let us define a disk as
    \begin{equation}
        C(c,r) \defeq \big\{ \lambda : \norm{c - \lambda} \le r \big\}.
    \end{equation}
    The theorem as presented in \citet{tretter_spectral_2008} shows that the eigenvalues of $\lambda(\A)$ are contained in a union of Gershgorin disks defined as follows
    \begin{equation}
        \lambda(\A)  \subset \bigcup^{L}_{i = 1} \Bigg\{ \bigcup\limits_{k=1}^{p_i - p_{i-1}} C\left( \lambda_k(\A_{ii}), \sum^{L}_{j=1,j\neq i}{ s_{max}(\A_{i,j})  } \right)
            \Bigg\}
    \end{equation}
    where the inner union is over a set disks for each eigenvalue of the block diagonal $\A_{i,i}$ while the outer union is over the L blocks in $\A$. The radius of the disk is constant for every eigenvalue in  the $i^{\mathrm{th}}$ diagonal block $\A_{i,i}$ and is given by the sum of singular values of the off diagonal blocks. Therefore, the largest eigenvalue of $\A$ lies in
        \begin{equation} \label{eq:maxGRSHG}
        \lambda_{max}(\A)  \subset \bigcup^{L}_{i = 1}  C\left( \lambda_{max}(\A_{ii}), \sum^{L}_{j=1,j\neq i}{ s_{max}(\A_{i,j})  } \right)
    \end{equation}
\subsection{Derivation of the expected singular values} \label{app:SingVals}
  \begin{align}
    \sigma_{max}\left({\G}_{\vec{\W^1}, \vec{\W^\beta}}\right)
    &= \EE\left[ \sigma_{max}\left(
    \phi(h) \ones \, \x^{0\TT}
    \right)\right]
    \otimes
    \EE\left[\sigma_{max}\left(
    \J^{\h^g\TT}_{\h^{1}}
    \mathbf{H}_g
    \J^{\h^g}_{\h^{\beta}}
    \right)
    \right]
        \\
    &= \sqrt{N^\beta} \left\vert  \EE\left[\phi(h)\right] \right\vert
    \left\lVert\EE\left[\x^0\right]\right\rVert_2
    \EE\left[  \sigma_{max}\left(\J^{\h^g\TT}_{\h^1}
    \mathbf{H}_g \J^{\h^g}_{\h^\beta}
    \right)
    \right]
  \end{align}
  \begin{equation}
    \le
    \sqrt{N^\beta} \left\vert  \EE\left[\phi(h)\right] \right\vert
    \left\lVert\EE\left[\x^0\right]\right\rVert_2
    \EE\left[  \sigma_{max}\left(\J^{\h^g\TT}_{\h^1}\right)
    \right]
    \EE\left[ \sigma_{max}\left(
     \mathbf{H}_g
     \right)
    \right]
    \EE\left[ \sigma_{max}\left(
    \J^{\h^g}_{\h^\beta}
    \right]
    \right)
  \end{equation}
  \begin{align}
      \sigma_{max}\left({\G}_{\vec{\W^1}, {b^\beta}}\right) &\le
      \EE\left[
      \sigma_{max}\left(
         \x^{0} {}\T
         \otimes \I
      \right)
      \right]
      \EE\left[
      \sigma_{max}\left(
      \J^{\h^g}_{\h^{\alpha}} {}\T
      \mathbf{H}_g
      \J^{\h^g}_{\h^{\beta}}
      \right)
      \right]\\
      &=
      \left\lVert\EE\left[\x^0\right]\right\rVert_2
      (\EE\left[\sigma_{max}\left. \J^{\h^g\TT}_{\h^1} \mathbf{H}_g \J^{\h^g}_{\h^\beta}
      \right)\right]\\
      & \le
      \left\lVert\EE\left[\x^0\right]\right\rVert_2
      \EE\left[  \sigma_{max}\left(\J^{\h^g\TT}_{\h^1}\right)
      \right]
      \EE\left[ \sigma_{max}\left(
       \mathbf{H}_g
       \right)
      \right]
      \EE\left[ \sigma_{max}\left(
      \J^{\h^g}_{\h^\beta}
      \right)
      \right]
    \end{align}
  \begin{align}
      \sigma_{max}\left({\G}_{b^1, \vec{\W^\beta}}\right) &\le
      \EE\left[
      \sigma_{max}\left(
         \x^{\beta - 1} {}\T
         \otimes \I
      \right)
      \right]
      \\
      &= \sqrt{N^\beta}
      \left\vert \EE\left[\phi(h)\right] \right\vert
      \EE\left[
        \sigma_{max}\left(
          \J^{\h^g\TT}_{\h^1} \mathbf{H}_g \J^{\h^g}_{\h^\beta}
        \right)
      \right]
  \end{align}
  \begin{equation}
    \le \left\vert \EE\left[\phi(h)\right] \right\vert
    \EE\left[  \sigma_{max}\left(\J^{\h^g\TT}_{\h^1}\right)
    \right]
    \EE\left[ \sigma_{max}\left(
     \mathbf{H}_g
     \right)
    \right]
    \EE\left[ \sigma_{max}\left(
    \J^{\h^g}_{\h^\beta}
    \right)
    \right]
  \end{equation}
  \begin{align}
      \sigma_{max}\left({\G}_{b^1, b^\beta}\right) &=
      \EE\left[
        \sigma_{max}\left(  \J^{\h^g\TT}_{\h^1} \mathbf{H}_g \J^{\h^g}_{\h^\beta}
        \right)
      \right]\\
      & \le
      \EE\left[  \sigma_{max}\left(\J^{\h^g\TT}_{\h^1}\right)
      \right]
      \EE\left[ \sigma_{max}\left(
       \mathbf{H}_g
       \right)
      \right]
      \EE\left[ \sigma_{max}\left(
      \J^{\h^g}_{\h^\beta}
      \right)
      \right]
  \end{align}

\subsection{Montecarlo estimate of spectral radius of $\mathbf{H}_g$ for 10 way softmax classification}\label{app:Hg}
  \begin{figure}[ht]
  	\centering
          \includegraphics[width=0.8\linewidth]{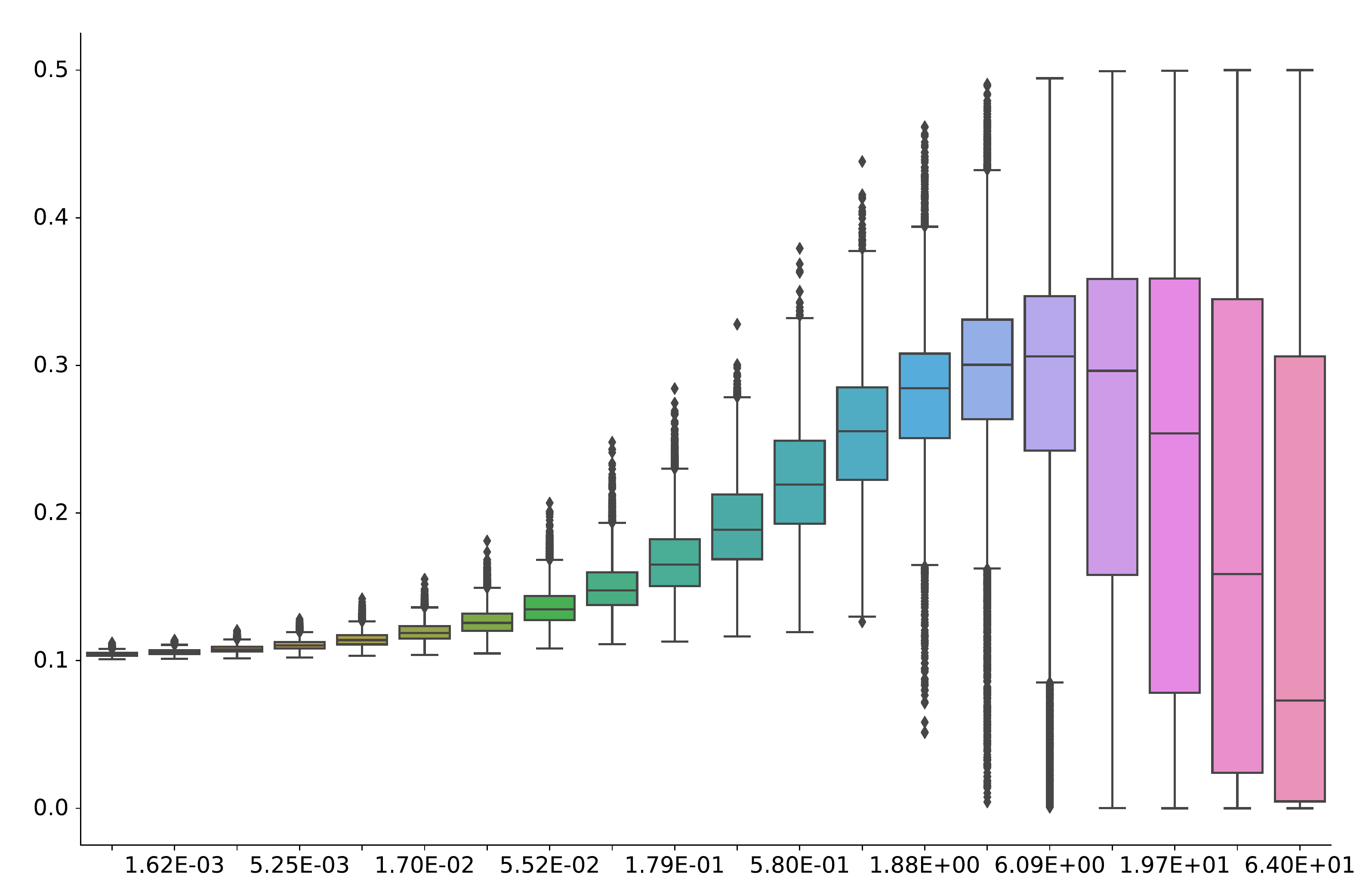}
      \caption{
      \textbf{ Distribution of $\lambda_{max}(\mathbf{H}_g)$ as a function of $q^*$}:
      In general, increasing the variance of the distribution of $h^g$ does not result in a monotonic increase in the spectral radius of the Hessian of the GLM layer. We plot the distribution of the maximum eigenvalues as a function of the variance of the softmax layer obtained from factorizing $10,000$ random matrices.
      }
  \end{figure}
\subsection{Manifold Optimization} \label{app:Manopt}
  The potentially non-convex constraint set constitutes a Riemannian manifold, when it is locally isomorphic to $\RR^n$, differentiable and endowed with a suitable (Riemannian) metric, which allows us to measure distances in the tangent space and consequentially also define distances on the manifold. There is considerable freedom in choosing a Riemannian metric; here we consider the metric inherited from the Euclidean embedding space which is defined as $\langle \W, \W^\prime \rangle \defeq \Tr(\W^{\prime\top}\W)$.
  To optimize a cost function with respect to parameters lying in a non-Euclidean manifold we must define a descent direction. This is done by defining a manifold equivalent of the directional derivative. An intuitive approach replaces the movement along a vector $\mathbf{t}$ with movement along a geodesic curve $\gamma(t)$, which lies in the manifold and connects two points $\W, \W^\prime \in \calM$ such that $\gamma(0) = \W, \, \gamma(1) = \W^\prime$. The derivative of an arbitrary smooth function $f(\gamma(t))$ with respect to $t$ then defines a tangent vector for each $t$.
  \paragraph{Tangent vector} $\xi_{\W}$ is a tangent vector at $\W$  if $\xi_\W$ satisfies $\gamma(0) = \W$ and
      \begin{equation}
       \xi_{\W} \defeq \frac{\mathrm{d}f(\gamma(t))}{\mathrm{d}t}\bigg\rvert_{t=0} \defeq \gamma^{\prime}(0) f
      \end{equation}
  The set of all tangents to $\calM$ at $\W$ is referred to as the tangent space to $\calM$ at $\W$ and is denoted by \tangent. The geodesic importantly is then specified by a constant velocity curve $\gamma^{\prime\prime}(t) = 0$ with initial velocity $\xi_\W$.
  To perform a gradient step, we must then move along $\xi_{\W}$ while respecting the manifold constraint. This is achieved by applying the exponential map defined as $\mathrm{Exp}_{\W}(\xi_\W) \defeq \gamma(1)$, which moves $\W$ to another point $\W^{\prime}$ along the geodesic.
  While certain manifolds, such as the Oblique manifold, have efficient closed-form exponential maps, for general Riemannian manifolds, the computation of the exponential map involves numerical solution to a non-linear ordinary differential equation \citep{absil_optimization_2007}. An efficient alternative to numerical integration is given by an orthogonal projection onto the manifold. This projection is formally referred to as a retraction $\mathrm{Rt}_\W : \tangent \to \calM$.

  Finally, gradient methods using Polyak (heavy ball) momentum (e.g. ADAM \citep{kingma_adam:_2014}) require the iterative updating of terms which naturally lie in the tangent space. The parallel translation $\calT_{\zeta}(\xi) : \mathit{T}\calM \bigoplus \mathit{T}\calM \to \mathit{T}\calM$ generalizes vector composition from Euclidean to non-Euclidean manifolds, by moving the tangent $\xi$ along the geodesic with initial velocity $\zeta \in \calT$ and endpoint $\W^\prime$, and then projecting the resulting vector onto the tangent space $\mathit{T}_{\W^\prime}\calM$. As with the exponential map, parallel transport $\calT$ may require the solution of non-linear ordinary differential equation. To alleviate the computational burden, we consider \textit{vector transport} as an effective, projection-like solution to the parallel translation problem. We overload the notation and also denote it as $\calT$, highlighting the similar role that the two mappings share.
  Technically, the geodesics and consequentially the exponential map, retraction as  well as transport $\calT$ depend on the choice of the Riemannian metric.
  Putting the equations together the updating scheme for Riemannian stochastic gradient descent on the manifold is
   \begin{equation}
     \W_{t+1} = \Pi_{\W_t}(-\eta_t \, \mathrm{grad}f)
   \end{equation}
      where $\Pi$ is either the exponential map $\mathrm{Exp}$ or the retraction $\mathrm{Rt}$ and $\mathrm{grad}f$ is the gradient of the function $f(\W)$ lying in the tangent space $\tangent$.
      \subsubsection{Optimizing over the Oblique manifold}\label{sec:optim:oblique}
      \citet{cho_riemannian_2017} proposed an updating scheme for optimizing neural networks where the weights of each layer are constrained to lie in the oblique manifold $\Ob(p,n)$. Using the fact that the manifold itself is a product of $p$ unit-norm spherical manifolds, they derived an efficient, closed-form Riemannian gradient descent updating scheme. In particular the optimization simplifies to the optimization over $\Ob(1,n)$ for each column  $\w_{i\in \{1,\ldots,p\}}$ of \W.
      \paragraph{Oblique gradient} The gradient $\mathrm{grad} f$ of the cost function f with 	respect to the weights lying in $\Ob(1,n)$ is given as a projection of the Euclidean gradient $\mathrm{Grad}f$onto the tangent at \w
       \begin{equation}
         \mathrm{grad}f = \mathrm{Grad}f - (\w {}\T \mathrm{Grad}f)\w
       \end{equation}
      \paragraph{Oblique exponential map} The exponential map $\Exp_\w$ moving $\w$ to $\w^{\prime}$ along a geodesic with initial velocity $\xi_\w$
       \begin{equation}
         \Exp_\w = \xi_\w \cos(\norm{\w}) + \frac{\w}{\norm{\w}}\sin(\norm{\w})
       \end{equation}
      \paragraph{Oblique parallel translation} The parallel translation $\calT$ moves the tangent vector $\xi_\w$ along the geodesic with initial velocity $\zeta_\w$
       \begin{align}
         \calT_{\zeta_{\w}}(\xi_\w) &=  \xi_\w - \\
              \frac{\zeta_{\w}}{\norm{\zeta_{\w}}}
          &(
              (1 - \cos(\norm{\zeta_{\w}}))
              + \w \sin(\norm{\zeta_{\w}})
              )
              \frac{\zeta_{\w}}{\norm{\zeta_{\w}}} {}\T
              \xi_\w
              \nonumber
       \end{align}

      \subsubsection{Optimizing over the Stiefel manifold}\label{sec:optim:stiefel}
      Optimization over Stiefel manifolds in the context of neural networks has been studied by \citep{harandi_generalized_2016,wisdom_full-capacity_2016,vorontsov_orthogonality_2017}. Unlike \cite{wisdom_full-capacity_2016,vorontsov_orthogonality_2017} we propose the parametrization using the Euclidean metric, which results in a different definition of vector transport.

      \paragraph{Stiefel gradient} The gradient $\mathrm{grad} f$ of the cost function f with 	respect to the weights lying in $\St(p,n)$ is given as a projection of the Euclidean gradient $\mathrm{Grad}f$onto the tangent at $\W$ \citep{edelman_geometry_1998,absil_optimization_2007}
       \begin{align} \label{eq:gradproj}
         \mathrm{grad}f &=
         (\mathbf{I} - \W\W {}\T)\mathrm{Grad}f
         \\
         &\quad + \frac{1}{2}\W \left(
              \W {}\T\mathrm{Grad}f - \mathrm{Grad}f {}\T\W
              \right)
              \nonumber
       \end{align}
      \paragraph{Stiefel retraction} The retraction $\Rt_\W(\xi_\W)$ for the Stiefel manifold is given by the Q factor of the QR decomposition \citep{absil_optimization_2007}.
       \begin{equation}
         \Rt_{\W}(\xi_\W) = \mathrm{qf}(\W + \xi_\W)
       \end{equation}
      \paragraph{Stiefel vector transport} The vector transport $\calT$ moves the tangent vector $\xi_\w$ along the geodesic with initial velocity $\zeta_\w$ for $\W \in \St(p,n)$ endowed with the Euclidean metric.
       \begin{equation}
         \calT_{\zeta_{\w}}(\xi_\w) =
             \left(\mathbf{I} - \mathbf{Y}\mathbf{Y} {}\T \right) \xi_\W + \frac{1}{2}\mathbf{Y}
              \left(
               \mathbf{Y} {}\T\xi_\W - \xi_\W {}\T\mathbf{Y}
              \right)
       \end{equation}
          where $\mathbf{Y} \defeq \Rt_\W(\zeta_\W)$. It is easy to see that the transport $\calT$ consists of a retraction of tangent $\zeta_\W$ followed by the orthogonal projection of $\eta_\W$ at $\Rt_\W(\zeta_\W)$. The projection is the same as the one mapping $\mathrm{P} : \mathrm{Grad}f \to \mathrm{grad}f $ in \eqref{eq:gradproj}.
  \subsubsection{Optimizing over non-compact manifolds}
    The critical weight initialization yielding a singular spectrum of the Jacobian tightly concentrating on $1$ implies that a substantial fraction of the pre-activations lie in expectation in the linear regime of the squashing nonlinearity and as a consequence the network acts quasi-linearly. To relax this constraint during training we allow the scales of the manifold constrained weights to vary. We chose to represent the weights as a product of a scaling diagonal matrix and a matrix belonging to the manifold. Then the optimization of each layer consists in the optimization of the two variables in the product. In this work we only consider isotropic scalings, but the method generalizes easily to the use of any invertible square matrix.

  \subsection{FIM and NTK have the same spectrum} \label{NTKJAC}
    The empirical Neural Tangent Kernel (NTK)
    Recall the definition in \eqref{eq:NTK}:
    \begin{equation}
      \hat{\Theta}_{t,i,j} \defeq \J^{\h^g}_{\x^0} \J^{\h^g \TT}_{\x^0}
    \end{equation}
    which gives a $N^g \vert \calD \vert$ by \(N^g \vert \calD \vert\) kernel matrix.
    By comparison the empirical Fisher Information matrix with a Gaussian likelihood is
    \begin{equation}
      \sum_{i=1^{|\calD|}} \J_{\theta}^{{h^{g}} \TT} \nabla^2_{h^g} \mathcal{L} \, \J_{\theta}^{h^g}
    \end{equation}
    To see that the spectra of these two coincide consider the third order tensor underlying both $\J^{\h^g}_{\h^1 i}$ for \(i \in 1 \dots \vert \mathcal{D} \vert \),
    additionally consider and unfolding \(\mathbf{A}\) with dimensions \(\vert \theta\vert \) by \(N^g \vert \mathcal{D}\vert\); i.e. we construct a matrix with dimension of number of parameters by number of outputs times number of data points. Then
      \begin{align}
        \bar{\G} = \mathbf{A}\T\mathbf{A}\\
        \hat{\Theta} = \mathbf{A}\mathbf{A}\T\\
      \end{align}
    and their spectra trivially coincide.\\
    \begin{remark}
        It is interesting to note that when the Fisher information metric and NTK are applied to a regression problem with Gaussian noise then the relation between admits the following interpretation.
          For \(\mathcal{L} = \frac{1}{2} \norm{\hat{y} - y}\)
          the Fisher information matrix \(\bar{\G}\) is the Riemannian metric on the tangent bundle and \(\hat{\Theta}\) is the Riemannian metric on the co-tangent bundle.
    \end{remark}

\end{document}

%% file: header.tex
\usepackage{times,wrapfig,amsmath,amsfonts,bm,color,enumitem,algorithm,algpseudocode}

\usepackage{times}
\usepackage{amsmath,amssymb,amsthm,bbm,mathtools}
\usepackage{graphicx}
\usepackage{subfigure}
\usepackage{hyperref}
\usepackage{url}
\usepackage{pbox}

\newcommand{\w}{\ensuremath{\mathbf{w}}}
\newcommand{\W}{\ensuremath{\mathbf{W}}}

\newcommand{\x}{\ensuremath{\mathbf{x}}}

\newcommand{\ones}{\mathbbm{1}}

\newcommand{\norm}[1]{\left\lVert{#1}\right\rVert}

\newcommand{\mnorm}[1]{\left\lVert{#1}\right\rVert} 

\newcommand{\defeq}{\triangleq}

\mathchardef\hyphen="2D

\usepackage{times}
\usepackage{amsmath,amssymb,amsthm}
\usepackage{graphicx}
\usepackage{subfigure}
\usepackage{hyperref}
\usepackage{url}
\usepackage{pbox}

\newcommand{\T}{^\top}
\newcommand{\TT}{\top }
\newcommand{\RR}{\mathbb{R}}
\newcommand{\EE}{\mathbb{E}}

\newcommand{\calD}{\mathcal{D}}

\newcommand{\calM}{\mathcal{M}}
\newcommand{\calN}{\mathcal{N}}

\newcommand{\calT}{\mathcal{T}}

\newcommand{\removed}[1]{}

\newcommand{\cov}{\mathrm{Cov}}

\renewcommand{\vec}[1]{\mathrm{vec}(#1)}

\newcommand{\St}{\ensuremath{\mathrm{St}}}
\newcommand{\Ob}{\ensuremath{\mathrm{Ob}}}
\newcommand{\I}{\ensuremath{\mathbf{I}}}
\newcommand{\Exp}{\ensuremath{\mathrm{Exp}}}
\newcommand{\Rt}{\ensuremath{\mathrm{Rt}}}
\newcommand{\tangent}{\ensuremath{\mathit{T}_{\mathbf{W}}\mathcal{M}}}
\DeclareMathOperator{\Tr}{Tr}
\newcommand{\h}{\mathbf{h}}
\newcommand{\J}{\mathbf{J}}
\newcommand{\A}{\mathbf{A}}
\newcommand{\G}{\mathbf{G}}
\newtheorem*{remark}{Remark}
\newtheorem{proposition}{Proposition}
\newtheorem{theorem}{Theorem}

\newtheorem*{definition}{Definition}
\newtheorem{lemma}{Lemma}
\newtheorem{asu}{Assumption}
\newcounter{subassumption}[asu]
\renewcommand{\thesubassumption}{(\textit{\roman{subassumption}})}
\makeatletter
\renewcommand{\p@subassumption}{\theasu}
\makeatother
\newcommand{\subasu}{
  \refstepcounter{subassumption}%
  \thesubassumption~\ignorespaces}

\def\eqref#1{equation~\ref{#1}}

